\pdfoutput=1
\documentclass[letterpaper, 10 pt, conference]{ieeeconf}
\IEEEoverridecommandlockouts
\usepackage{amsmath}
\usepackage{amssymb}
\usepackage{booktabs}
\usepackage{nicefrac}
\usepackage{graphicx}
\usepackage{caption}
\usepackage[font=footnotesize]{subcaption}
\usepackage{siunitx}
\usepackage{comment}
\usepackage{caption}
\usepackage{pgf}
\usepackage{xcolor}
\usepackage{placeins}
\usepackage{dblfloatfix}
\usepackage{multicol}
\usepackage{bm}
\usepackage{dsfont}
\usepackage[normalem]{ulem}
\usepackage{flushend}
\usepackage{balance}

\makeatletter             %
\let\NAT@parse\undefined  %
\makeatother              %
\usepackage[colorlinks=true, allcolors=blue]{hyperref}

\usepackage[capitalise]{cleveref}

\newcommand{\diff}[1]{{\mathrm{d}#1}}

\newcommand{\E}[1]{\operatorname{\mathbb{E}}[ #1 ]}
\newcommand{\R}[0]{\operatorname{\mathbb{R}}}
\newcommand{\SO}[1]{{\mathrm{SO}(#1)}}
\newcommand{\SE}[1]{{\mathrm{SE}(#1)}}

\newcommand{\norm}[1]{\lVert #1 \rVert}

\title{\LARGE \bf
  Joint Localization and Planning using Diffusion
}

\author{Lukas Lao Beyer$^{1}$ and Sertac Karaman$^{1}$%
  \thanks{$^{1}$Laboratory for Information and Decision Systems, Massachusetts
    Institute of Technology, Cambridge, MA. \texttt{\{llb, sertac\}@mit.edu}}%
}

\begin{document}
\maketitle
\thispagestyle{empty}
\pagestyle{empty}

\begin{abstract}
  Diffusion models have been successfully applied to robotics problems
  such as manipulation and vehicle path planning. In this work, we
  explore their application to end-to-end navigation -- including both
  perception and planning -- by considering the problem of jointly
  performing global localization and path planning in known but
  arbitrary 2D environments. In particular, we introduce a diffusion
  model which produces collision-free paths in a global reference
  frame given an egocentric LIDAR scan, an arbitrary map, and a
  desired goal position. To this end, we implement diffusion in the
  space of paths in $\SE{2}$, and describe how to condition the
  denoising process on both obstacles and sensor observations. In our
  evaluation, we show that the proposed conditioning techniques enable
  generalization to realistic maps of considerably different
  appearance than the training environment, demonstrate our model's
  ability to accurately describe ambiguous solutions, and run
  extensive simulation experiments showcasing our model's use as a
  real-time, end-to-end localization and planning stack.
\end{abstract}

\section{Introduction}

\begin{figure}[t]
  \centering
  \includegraphics[width=\linewidth]{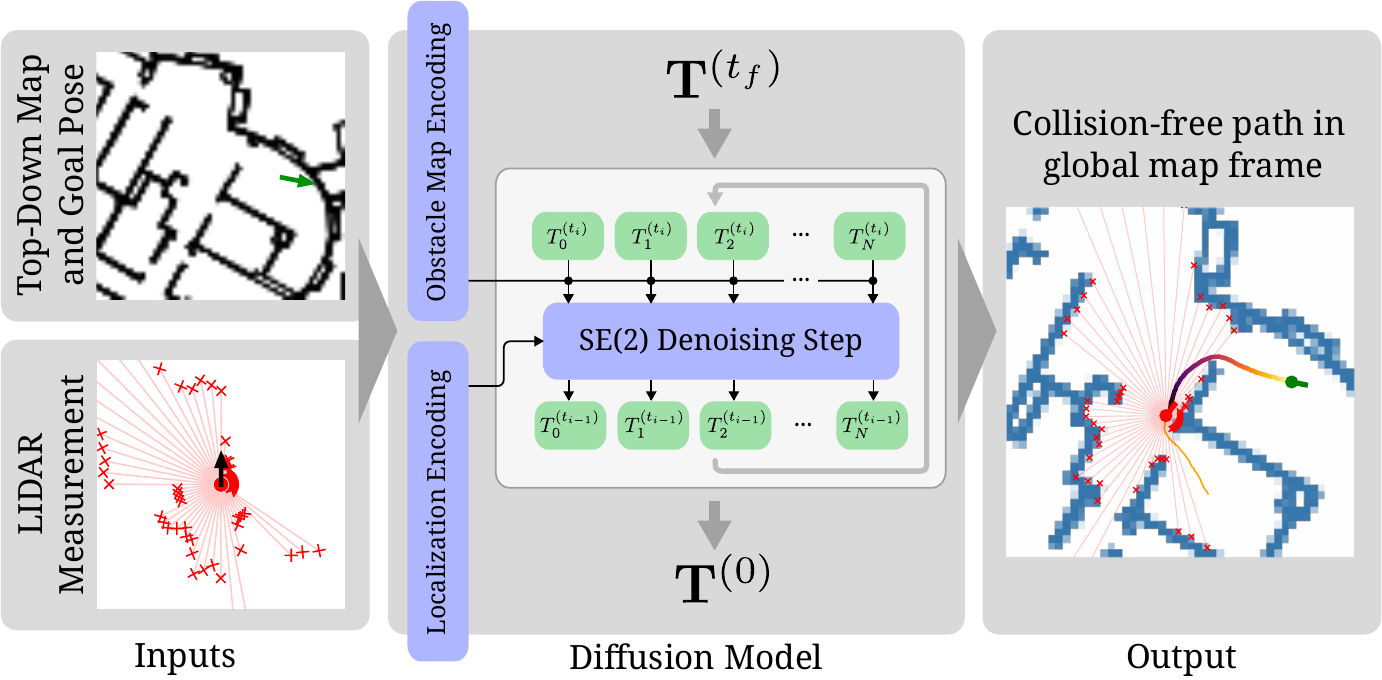}
  \caption{Proposed model. A denoising diffusion process is
    conditioned on an obstacle map, a LIDAR scan, and a goal pose,
    producing a collision free path in the global map frame.}
  \label{fig:overview}
\end{figure}

Denoising diffusion probabilistic models~\cite{ho2020denoising} have
shown to be a powerful tool for sampling from complicated,
high-dimensional distributions, achieving state-of-the-art performance
on tasks such as image generation~\cite{dhariwal2021diffusion}, motion
planning~\cite{janner2022planning} and
control~\cite{chi2024diffusionpolicy}.
In this paper, we introduce a diffusion model that can solve a vehicle
navigation task consisting of localization and planning in arbitrary
2D environments. In particular, our model is conditioned on a 2D
obstacle map, raw LIDAR sensor measurements, and a desired goal state,
and produces collision-free paths in the global map frame (see
\cref{fig:overview}). We also demonstrate how this model's output can
serve to control a vehicle with real-time online replanning. To the
best of our knowledge, this is the first paper exploring the joint
global vehicle localization and planning problem using
diffusion. However, there is a significant amount of existing work on
applications of diffusion to several problems in robotics.

\textbf{Diffusion planning and RL.}
\textit{Diffuser}~\cite{janner2022planning} uses a diffusion model in
conjunction with a guidance function learned via reinforcement
learning (RL) to perform a variety of planning tasks. The use of
hand-designed guidance to enforce test-time conditions, such as
obstacle avoidance, in diffusion models for
planning~\cite{song2023loss}, has also been explored. Offline RL has
been shown to benefit from diffusion models to represent
policies~\cite{wang2022diffusion}, and conditional diffusion models
have been used to behavior-clone a model-based 2D
pathplanner~\cite{liu2024dipper}.

We note that these contributions generally do not consider the
geometry of the diffused states specially, performing diffusion in
Euclidean space, and do not address the global localization problem,
requiring an external perception and control pipeline. In contrast,
\textit{Diffusion Policy} learns visumotor policies that directly take
sequences of images as inputs~\cite{chi2024diffusionpolicy} with
impressive results in manipulation applications. However,
\textit{Diffusion Policy} has not been applied to problems related to
vehicle navigation.

\textbf{Diffusion on manifolds.} Diffusion on~$\SE{3}$ has been
applied to manipulation problems~\cite{urain2023se3}. Recent work on
diffusion on Riemannian manifolds~\cite{de2022riemannian} has also
laid the groundwork for the rigorous development of diffusion
on~$\SO{3}$~\cite{leach2022denoising} and $\SE{3}$ with applications
to protein design~\cite{yim2023se3}. These results transfer directly
to applications in robotic navigation.

\textbf{Perception using diffusion.} Diffusion for LIDAR localization
has been studied~\cite{li2024diffloc} in the context of absolute pose
regression with a given map known at training time. The LIDAR
localization problem we consider in our work differs from this in that
we do not rely on per-map model training but instead can condition on
arbitrary maps at test time. This is more similar to the point cloud
registration studied using diffusion by
Wu~et~al.~\cite{wu2023pcrdiffusion}, although the global localization
problem differs in that the map may describe a much larger extent than
captured by the sensor observation.

\textbf{End-to-end navigation.} Prior work on end-to-end
navigation~\cite{liu2021efficient, amini2019variational} explores a
similar problem setting, but uses explicit representations of
distributions such as Gaussian mixture models to handle
uncertainty. Diffusion models have the potential to characterize much
richer distributions, which we demonstrate in our experiments.

In summary, we present the following main contributions in this
paper. First, obstacle-free trajectory generation via diffusion on
$\SE{2}$, conditioned on arbitrary obstacle maps. Second, a
conditioning technique enabling our diffusion model to perform global
localization given an arbitrary map and egocentric LIDAR
scan. Finally, a demonstration of jointly solving the global
localization and planning tasks using a diffusion model and the use of
our model for closed-loop control in realistic environments.

\section{Preliminaries}

In this work, we focus on vehicles traversing 2D
environments. Therefore, we consider paths parameterized by $N$
approximately uniformly spaced pose samples~${\mathbf{T} = [T_1,
    \dots, T_N]} \in \SE{2}^N$, where each pose~${T \in \SE{2}}$
consists of a heading~${R \in \SO{2}}$ as well as a position~$X \in
\R^2$ in the global coordinate frame. We use similar notation
$\mathbf{R} \in \SO{2}^N$ and $\mathbf{X} \in \R^{2N}$ for sequences
of rotations and translations, respectively. We also assume that
positions are scaled such that their coordinates do not fall far
outside of the~$[-1, 1]$ range.

\subsection{Forward and Reverse Diffusion Processes on $\SE{2}$}

\newcommand{\fwd}[1]{\mathbf{#1}^{(t)}}
\newcommand{\rev}[1]{\overleftarrow{\mathbf{#1}}^{(t)}}

To define a forward and reverse diffusion process on~$\SE{2}$, we
follow the development of diffusion modeling on~$\SE{3}$ by Yim et
al.~\cite{yim2023se3}. In particular, we leverage the fact that
$\SE{2}$ can be identified with $\SO{2} \times \R^2$ in order to
define a forward process~$(\mathbf{T}^{(t)})_{t \geq 0}$ on~$\SE{2}$
by considering diffusion on~$\SO{2}$ and~$\R^2$
separately:
\begin{equation}
  \diff{\fwd{R}} = g(t) \diff{\mathbf{B}^{(t)}_{\SO{2}}}\,\,\text{and}\,\,
  \diff{\fwd{X}} = g(t) \diff{\mathbf{B}^{(t)}_{\R^2}}.
  \label{eq:forward-sde}
\end{equation}
Here, $g(t)$ is the diffusion coefficient, and
$\diff{\mathbf{B}^{(t)}_{\SO{2}}}$ and
$\diff{\mathbf{B}^{(t)}_{\R^2}}$ denote Brownian motion on $\SO{2}$
and $\R^2$, respectively. As in Karras et al.'s EDM
model~\cite{karras2022elucidating}, we choose to skip the drift term.

Let $\rev{T} = \mathbf{T}^{(t_f - t)}$, where $t_f$ denotes the final
timestep of the forward diffusion process. Define $\rev{R}$ and
$\rev{X}$ equivalently. Let $p_t$ denote the density of
$\mathbf{T}^{(t)}$. Then, following {Song et
  al. \cite{song2021scorebased}} and {De Bortoili et
  al. \cite{de2022riemannian}}, the time reversal of the forward
process~\eqref{eq:forward-sde} is given by
\begin{align}
\begin{split}
  \diff{\rev{T}}
  &= g^2(t_f - t) \nabla_{\rev{T}} \log p_{t_f-t}(\rev{T}) \\
  &\quad+ g(t_f - t) [\diff{\mathbf{B}_{\SO{2}}^{(t)}}, \diff{\mathbf{B}_{\R^2}^{(t)}}]
  \label{eq:reverse}
\end{split}
\end{align}
so that for $t \in [0, t_f]$, we have $\rev{T} \sim p_{t_f - t}$.

\subsection{Score Modeling on $\SE{2}$}

To sample from the data distribution $p_0$ by running reverse
diffusion, we approximate the intractable Stein score~${\nabla \log
  p_t}$. Using denoising score matching (DSM), a neural network
$s_\theta(t, \cdot)$ is trained to minimizing the DSM loss
\begin{equation}
  \mathcal{L}(\theta) = \E{\lambda_t \norm{
      \nabla \log p_{t|0}(\mathbf{T}^{(t)} \mid \mathbf{T}^{(0)})
      - s_\theta(t, \mathbf{T}^{(t)})
    }^2}
  \label{eq:dsm}
\end{equation}
with weights $\lambda_t > 0$ and $t \in [0, t_f]$~\cite{song2021scorebased}.
Since we designed the diffusion processes on~$\SO{2}$ and~$\R^2$ to be
independent~\eqref{eq:forward-sde}, note that the conditional score
\begin{equation}
  \begin{aligned}
  \nabla \log p_{t|0}(\mathbf{T}^{(t)} \mid \mathbf{T}^{(0)})
  = [&\nabla_{\mathbf{R}^{(t)}} \log p_{t|0}(\mathbf{R}^{(t)} \mid \mathbf{R}^{(0)}), \\
    &\nabla_{\mathbf{X}^{(t)}} \log p_{t|0}(\mathbf{X}^{(t)} \mid \mathbf{X}^{(0)})]
  \label{eq:score}
  \end{aligned}
\end{equation}
can be computed by considering the rotation and translation
separately~\cite{yim2023se3}. Here, we have for the Euclidean part of
the score that $\nabla_{\mathbf{x}} \log p_{t|0}(\mathbf{x} \mid
\mathbf{y}) = \sigma^{-2}(t) (\mathbf{y} - \mathbf{x})$, where we
refer to Karras et al.'s EDM formulation~\cite{karras2022elucidating}
for the definition of $\sigma(t)$ in terms of the diffusion
coefficient~$g(t)$. In the case of angles $\phi, \psi \in \SO{2}$, the
score is instead computed by differentiating the wrapped
normal~\cite{fletcher2003gaussian} pdf:
\begin{equation}
  \nabla_{\phi} \log p_{t|0}(\phi \mid \psi)
  = \nabla_{\phi} \log \sum_{k\in\mathbb{N}}
  \exp(-\frac{(\psi - \phi - 2 \pi k)^2}{2 \sigma^2(t)}).
\end{equation}
In practice we observe that the series converges rapidly
on~${[-\pi,\pi)}$, so we truncate it summing only
over~${k\in[-10,10]}$ and compute the derivative using automatic
differentiation, or use the Euclidean score as an approximation in
the case where $\sigma(t)$ is small.

\section{Diffusion Localization and Planning Model}

We explore a diffusion model for jointly performing global
localization and planning in the context of behavior cloning of a
model-based pathplanner. We procedurally generate a
dataset~$\mathcal{D} = \{\mathcal{S}_i\}_{i \in \mathbb{Z}}$ of
example scenarios and demonstrations.  Each scenario $\mathcal{S} =
(\mathcal{E}, \mathcal{O}, \mathcal{G}, \mathbf{T}^*)$ consists of a
randomly generated environment occupancy map~$\mathcal{E} \in \{0,
1\}^{H \times W}$, an noisy egocentric LIDAR sensor
observation~$\mathcal{O} \in \R^{N_{\text{rays}}}$, a goal pose
$\mathcal{G} \in \SE{2}$, and an expert demonstration produced by a
model-based pathplanner in the form of a collision free
path~$\mathbf{T}^* \in \SE{2}^N$.

\subsection{Denoising Network}

We describe our score approximator in terms of a
``denoiser''~${f_\theta(t, \cdot) : \SE{2}^N \rightarrow \SE{2}^N}$,
as follows:
\begin{equation}
  s_\theta(t, \mathbf{T}^{(t)})
  = \nabla \log p_{t|0}(\mathbf{T}^{(t)} \mid f_\theta(t, \mathbf{T}^{(t)})).
\end{equation}
Note that we have omitted writing explicit dependencies on
$\mathcal{E}$, $\mathcal{O}$ and $\mathcal{G}$ for notational
simplicity. However, $f_\theta$ and the entire diffusion processes are
to be understood as being conditioned on $\mathcal{E}$, $\mathcal{O}$
and $\mathcal{G}$ as applicable.
More explicitly, we write $f_\theta(t, \cdot)$ in terms of the
conditional 1D
U-Net~\cite{ronneberger2015u}~$F_{\theta}(\cdot~\mid~\mathbf{x}_{\text{cond}}):~\R^{N
  \times (4 + d_{\text{local}})} \rightarrow \R^{N \times 4}$ as
\begin{equation}
  f_\theta(t, \mathbf{T}^{(t)})
  = f_{\text{out}}(F_\theta(f_{\text{in}}(\mathbf{T}^{(t)}, \mathcal{E})
            \mid f_{\text{cond}}(t, \mathcal{O}, \mathcal{G}))).
\end{equation}
Here, $f_{\text{in}}(\cdot, \mathcal{E}) : \SE{2}^N \rightarrow \R^{N
  \times (4 + d_{\text{local}})}$ encodes the position $(x, y)$ and
rotation angle $\phi$ of each input pose $T_i^{(t)}$ as a vector
$\begin{bmatrix} x & y & \cos \phi & \sin \phi \end{bmatrix}^\top$ and
concatenates a $d_{\text{local}}$-dimensional \textit{local}
conditioning vector to each encoded input pose. An additional
\textit{global} conditioning vector is computed by $f_{\text{cond}}$
and applied via FiLM modulation~\cite{perez2018film}. This global
conditioning vector always includes the goal pose and a sinusoidal
positional embedding of the current timestep $t$. Finally,
$f_{\text{out}} : \R^{N \times 4} \rightarrow \SE{2}^N$ undoes the pose
transformation and encoding to recover the denoised path.

\subsection{Obstacle Avoidance} \label{sec:obstacle-conditioning}

\begin{figure}[tb]
  \centering
  \includegraphics[trim=2.7cm 3cm 2.5cm 2cm, width=0.49\linewidth]{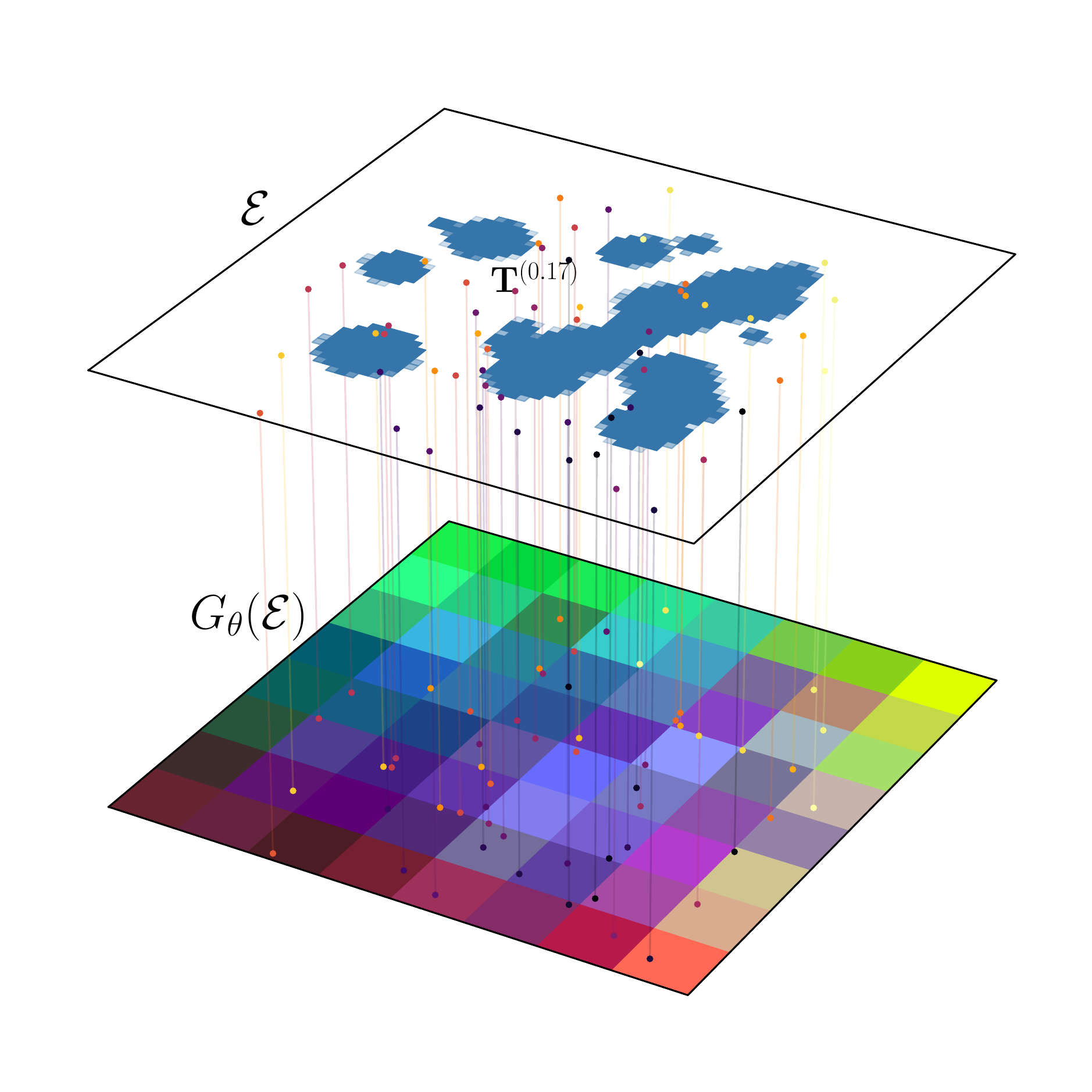}
  \includegraphics[trim=2.7cm 3cm 2.5cm 2cm, width=0.49\linewidth]{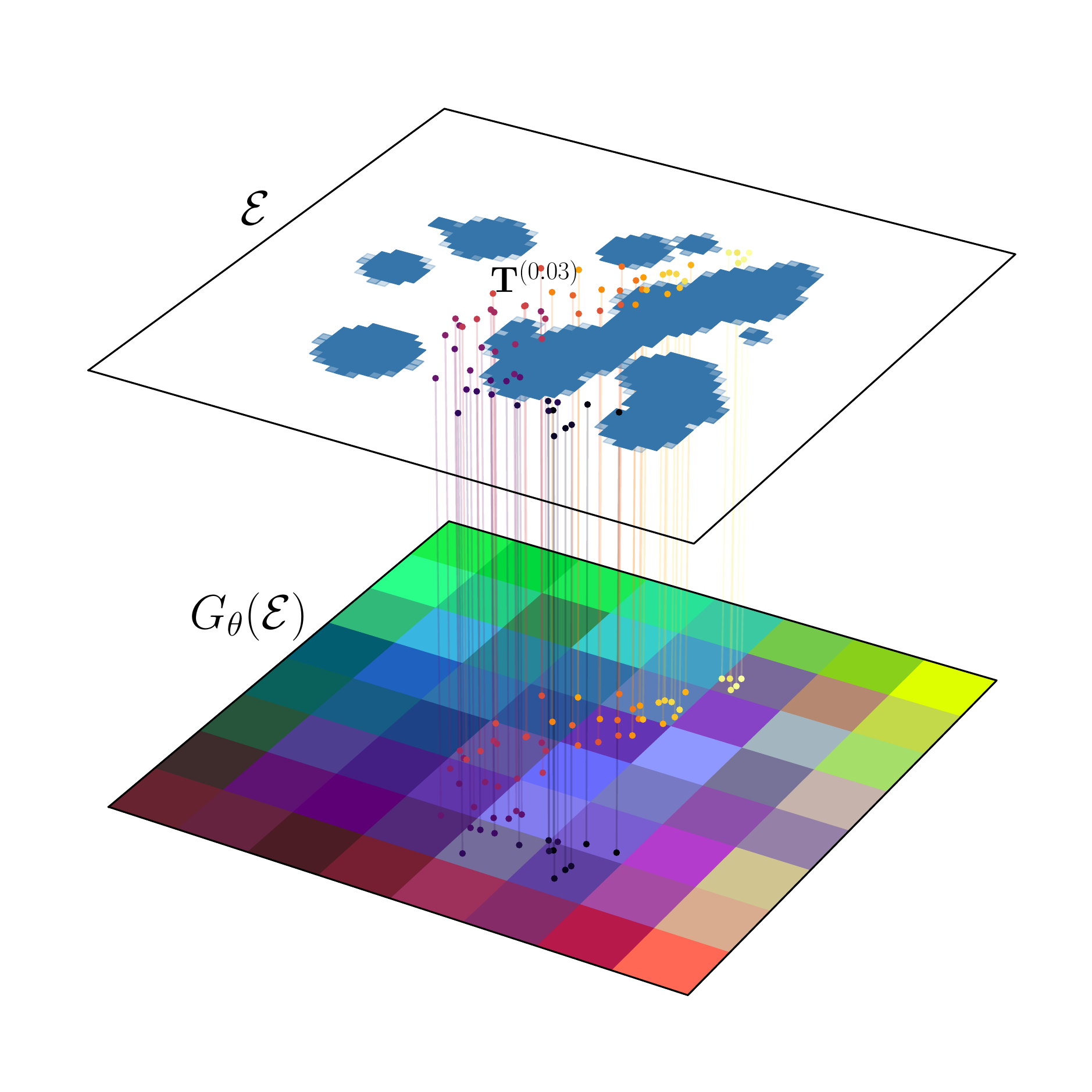}
  \caption{Local conditioning strategy based on sampling of the
    encoded obstacle map~$G_\theta(\mathcal{E})$ shown for two
    different noise levels. Samples of~$G_\theta(\mathcal{E})$ are
    appended to the corresponding pose and fed into the denoising
    network.}
  \label{fig:env-feature-map-sampling}
\end{figure}

We propose a simple local conditioning strategy to condition the
denoising network on the obstacle map~$\mathcal{E}$. To this end, we
first encode~$\mathcal{E}$ into a feature map via a learned encoder
network $G_\theta: \R^{H \times W} \rightarrow \R^{d_{\text{local}}
  \times H' \times W'}$. The encoded map~$G_\theta(\mathcal{E})$ is
then sampled (via bilinear interpolation) at the
positions~$\mathbf{X}^{(t)}$ corresponding to each pose in the (noisy)
input path, and $f_\text{in}$ concatenates each sampled feature to the
corresponding pose encoding. Sampling the encoded map at out-of-bounds
locations produces a zero feature
vector. \cref{fig:env-feature-map-sampling} illustrates this sampling
process.

\begin{figure}[tb]
  \centering
  \includegraphics[width=0.235\linewidth]{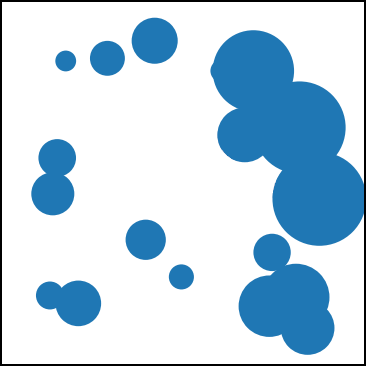}
  \includegraphics[width=0.235\linewidth]{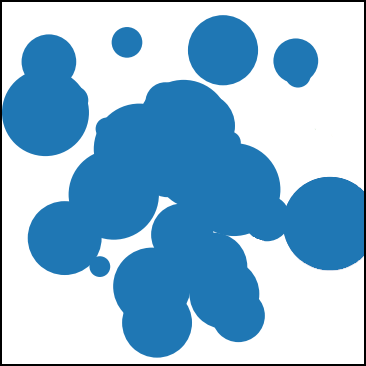}
  \includegraphics[width=0.235\linewidth]{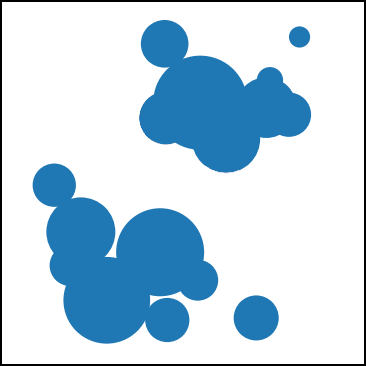}
  \includegraphics[width=0.235\linewidth]{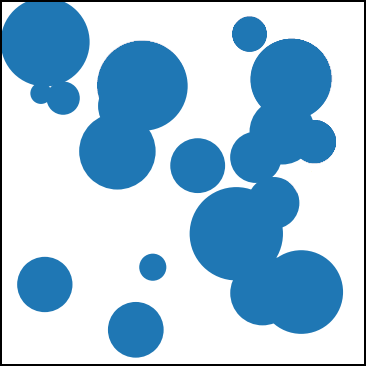}%
  \vspace{3pt}\\
  \includegraphics[width=0.235\linewidth]{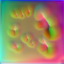}
  \includegraphics[width=0.235\linewidth]{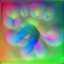}
  \includegraphics[width=0.235\linewidth]{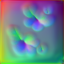}
  \includegraphics[width=0.235\linewidth]{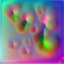}
  \caption{Obstacle map encoding using U-Net encoder. Top row shows
    test environments with obstacles in blue. Bottom row visualizes
    the corresponding encoded obstacle feature maps by mapping the
    first three principal components of each feature onto the RGB
    channels. Feature maps contain structure reminiscent of a Voronoi
    decomposition and also appear to encode distance to obstacles. }
  \label{fig:env-feature-maps}
\end{figure}

Intuitively, for the model to perform obstacle avoidance successfully,
the map encoder must learn to produce features which capture
information which is locally relevant for the obstacle avoidance task
while incorporating global geometric knowledge of the full obstacle
map. In \cref{fig:env-feature-maps} we empirically observe such
behavior.

Note also that the map encoder~$G_\theta$ does not depend on~$t$, so
at test time, the encoded map~$G_\theta(\mathcal{E})$ is reused across
iterations during the reverse diffusion process.

\subsection{Global Localization} \label{sec:observation-conditioning}

\begin{figure}[tb]
  \centering
  \includegraphics[trim=2.8cm 3cm 2.5cm 1.5cm, width=0.478\linewidth]{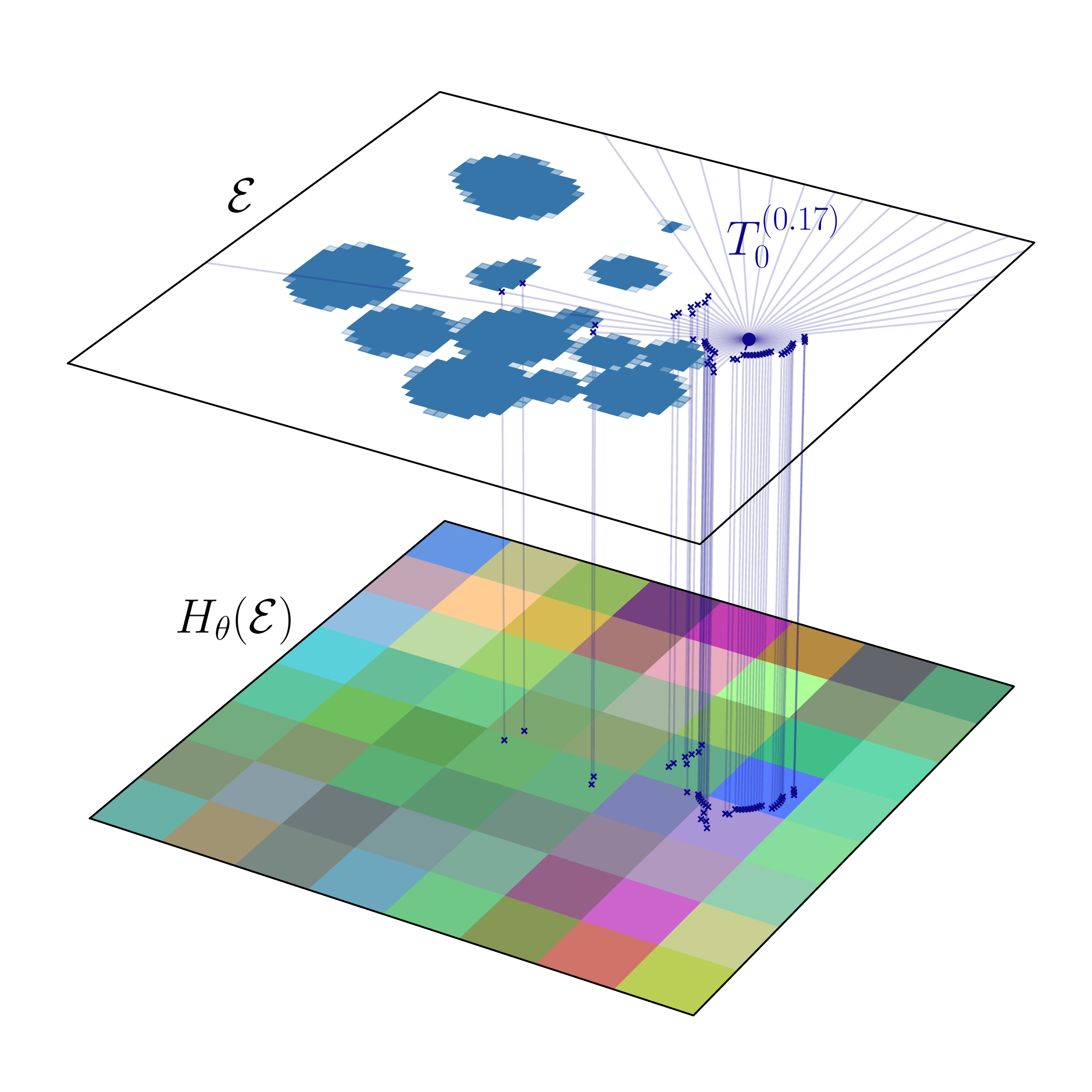}
  \includegraphics[trim=2.8cm 3cm 2.5cm 1.5cm, width=0.478\linewidth]{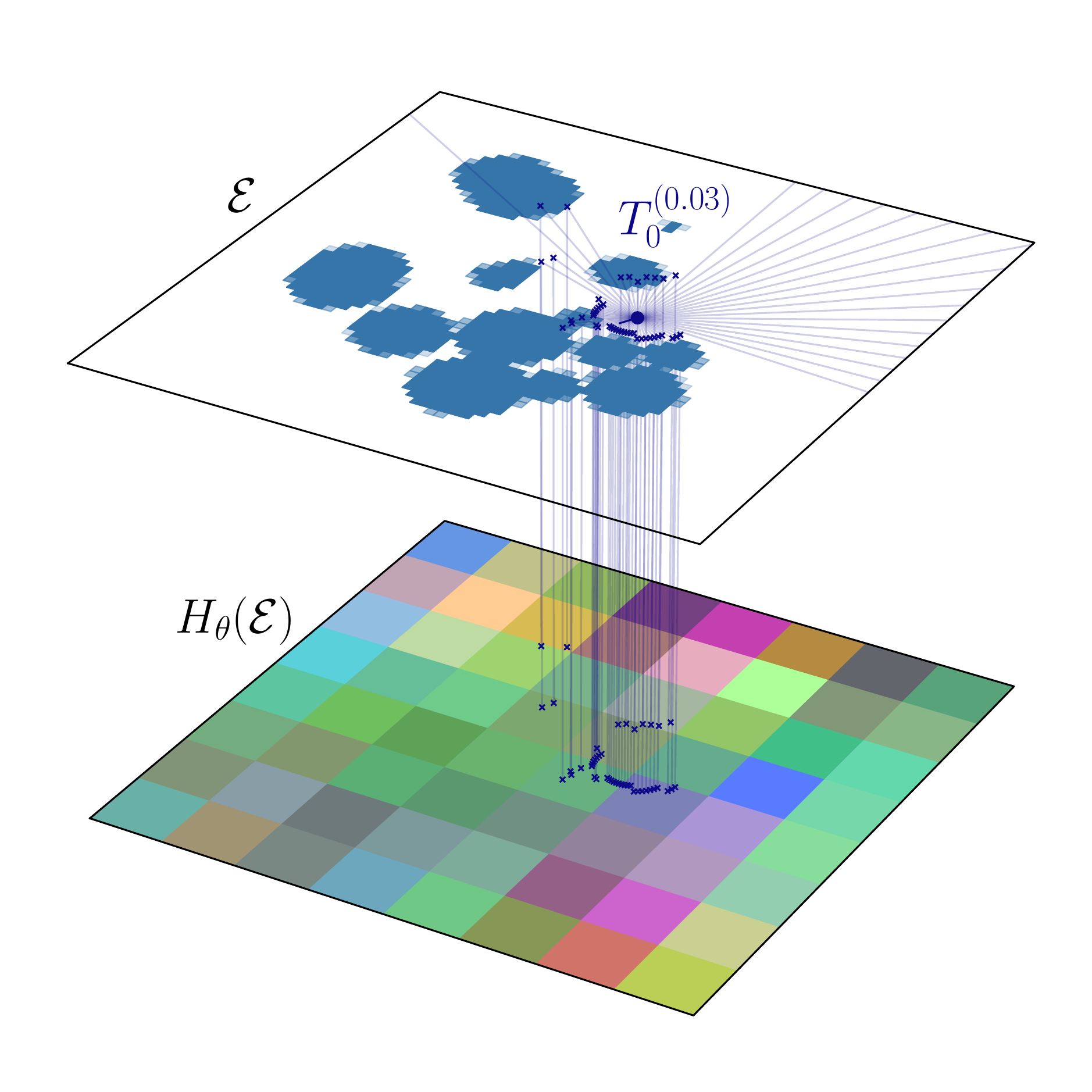}
  \caption{Sensor observation conditioning for global
    localization. Given the (noisy) start pose~$T_0^{(t)}$ and LIDAR
    observation~$\mathcal{O}$, we calculate the termination position
    of each ray to determine the location at which to sample the
    localization feature map~$H_\theta(\mathcal{E})$. The
    concatenation of the sampled features serves as conditioning for
    the denoising network. }
  \label{fig:loc-feature-sampling}
\end{figure}

Similar to the local conditioning strategy for obstacle avoidance from
\cref{sec:obstacle-conditioning}, we introduce an environment map
encoder $H_\theta: \R^{H \times W} \rightarrow \R^{d_{\text{loc}}
  \times H' \times W'}$ and again adopt a conditioning method based on
sampling the feature map~$H_\theta(\mathcal{E})$. In this case, we
sample the feature map at the termination position of each LIDAR ray,
assuming the (noisy) start pose of the trajectory,~$T_1^{(t)}$, as the
reference frame. This yields $N_\text{rays}$ features, which are
concatenated and fed to the denoising U-Net via FiLM
conditioning. This process is illustrated
in~\cref{fig:loc-feature-sampling}.

\subsection{Joint Localization and Planning} \label{sec:joint-loc-plan}

We train $F_\theta$, $G_\theta$, and $H_\theta$ jointly by minimizing
a score matching loss~\eqref{eq:dsm}. Weights $\lambda_t$, noise
schedule~$\sigma(t)$, and distribution of samples of $t$ during
training are chosen following the EDM framework of Karras et
al.~\cite{karras2022elucidating}.

At test time, the model is given a novel environment
map~$\mathcal{E}$, a sensor observation~$\mathcal{O}$, and a goal pose
in the global map frame~$\mathcal{G}$. Sampling from the reverse
diffusion process then produces an estimated path in the global map
frame, ideally starting from a correct estimate of the current
location according to the LIDAR measurements and traversing the
environment towards the specified goal.

As in \textit{Diffuser}~\cite{janner2022planning}, we additionally
implement an incremental sampling strategy which leverages previously
generated plans to warm-start the diffusion process. In an online
replanning setting, this allows us to apply a small amount of noise to
the previous path, requiring a much smaller number of denoising
iterations to replan with updated observations. We highlight that this
approach also prevents ``mode confusion'': from-scratch planning in
each frame can lead to samples coming from different modes of the
distribution of plans, while a warm start serves as a form of
conditioning on the previous solution that we observe to prevent
unnecessary mode switches.

\section{Dataset Generation} \label{sec:dataset-generation}

Our training dataset consists of smooth, obstacle-free paths
traversing cluttered 2D environments between randomized start and goal
positions. Each example scenario is generated by placing a variable
number of circular obstacles of randomized position and radius. We
render the obstacle map to a~$64~\times~64$ pixel bitmap serving as
the environment map~$\mathcal{E}$. The vector of LIDAR ray
lengths~$\mathcal{O}$ is computed by casting 64 rays in the rendered
environment map~$\mathcal{E}$, starting from the ground truth start
position until hitting an obstacle.

The reference path~$\mathbf{T}^*$ is generated in three steps:
shortest path search, spline fitting and optimization, and heading
assignment. First, an~A* search attempts to find the shortest
obstacle-free path between start and goal on a discrete grid. If the
search succeeds, the second step fits a 2D cubic B-spline to the A*
path, and then optimizes it considering obstacle avoidance and and
length minimization costs. This step is implemented as a nonlinear
optimizations using the Ceres~\cite{Agarwal_Ceres_Solver_2022}
library.

\begin{figure}
  \centering
  \vspace{4pt}
  \includegraphics[width=0.32\linewidth]{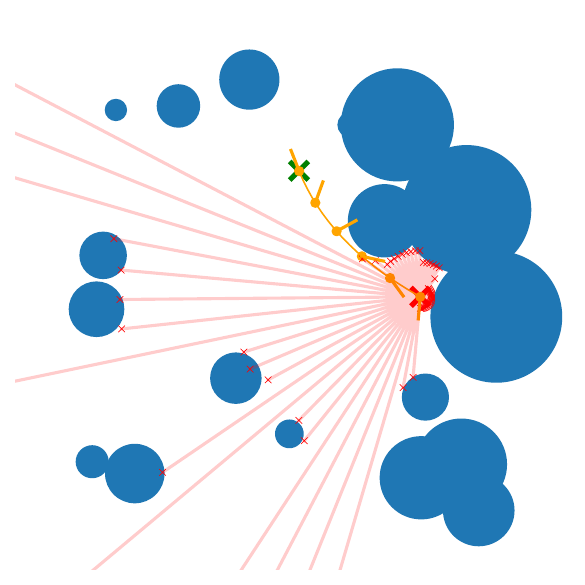}
  \includegraphics[width=0.32\linewidth]{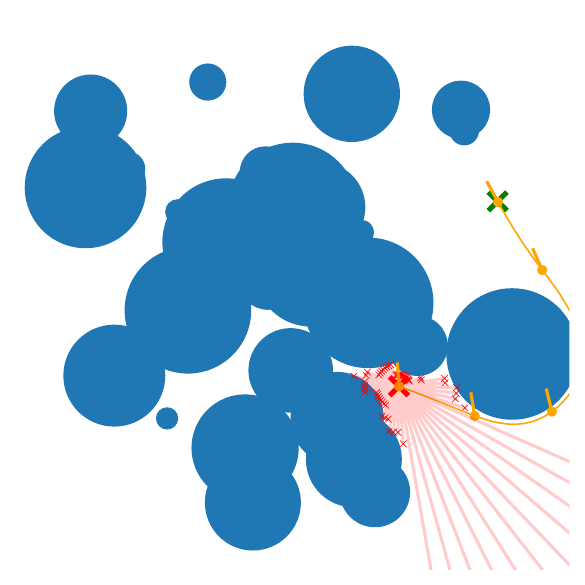}
  \includegraphics[width=0.32\linewidth]{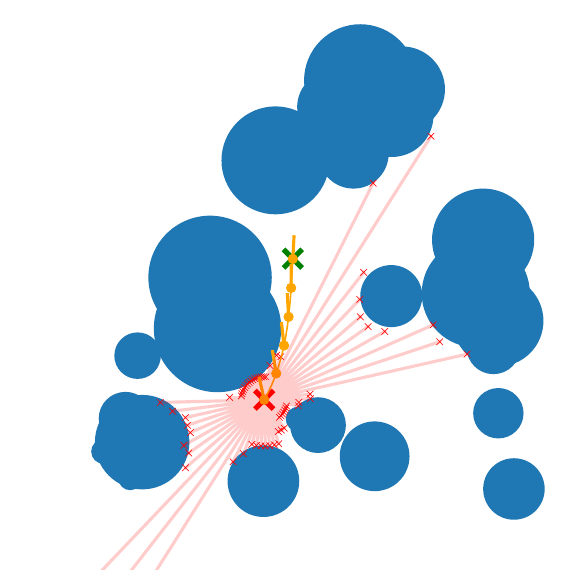}%
  \vspace{-2pt}
  \caption{Random example scenarios produced by dataset generation
    procedure. Obstacles shown in blue, expert trajectory produced by
    B-spline optimization shown in orange, and LIDAR scan shown in
    red.}
  \label{fig:example-scenarios}
\end{figure}

In a final step, we assign the heading along the path by randomly
choosing a start heading and linearly (as a function of arclength)
interpolating it towards the tangent heading at the goal position. We
also perform collision checks of the final optimized spline and
discard the scenario if the optimization in the third step produced an
invalid result. \cref{fig:example-scenarios} shows a random selection of
scenarios produced by the procedure described in this section.

\section{Implementation and Evaluation}

We evaluate our model first on the pure localization task to verify
the effectiveness of the proposed conditioning technique, assessing
global localization accuracy, generalization to out-of-distribution
environments, and distributional modeling capability. We then
demonstrate our model on the full navigation task consisting of joint
global localization and planning, starting with a quantitative
evaluation of success rate on our synthetic dataset and ending with an
application to real-time closed-loop control in a realistic scenario.
All reported runtimes are measured
on an NVIDIA RTX A5000 GPU, and models are trained on two RTX A6000
GPUs for one week.

\subsection{Implementation Details}

The first three layers of a ResNet-18~\cite{he2016deep} network are
used as the environment and localization encoders~$G_\theta$
and~$H_\theta$, producing~$8 \times 8$ feature maps from the $64
\times 64$ input obstacle maps. To generate the high resolution
feature maps shown in \cref{fig:env-feature-maps}, we train a model
using a larger U-Net as the environment encoder, but find it does not
significantly improve planning or localization performance. The pose
denoising network~$F_\theta$ is a 1D U-Net with three down-/upsampling
and stages and four ResNet blocks in each stage.

\subsection{Global Localization}

Next, we evaluate the quality of a diffusion-based global localization
model. This model leverages the LIDAR observation conditioning
described in~\cref{sec:observation-conditioning} to condition a
denoising multilayer perceptron with 6 layers of size 1024 (instead of
the 1D U-Net used for path diffusion) to estimate the pose of the
sensor in the global map frame. Conditioning is performed by
concatenation to the input pose.

\begin{figure}[tb]
  \centering
  \includegraphics[width=\linewidth]{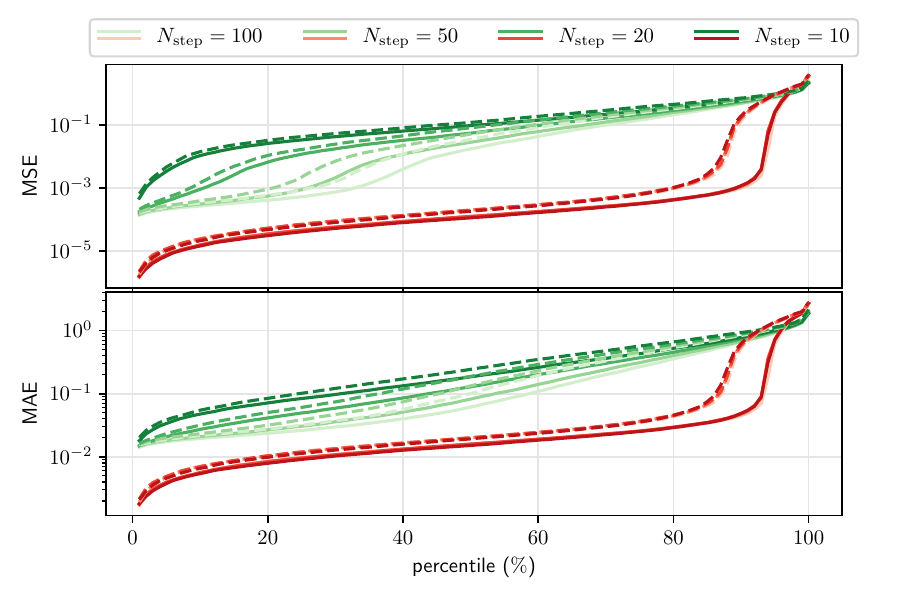}%
  \vspace{-7pt}
  \caption{Localization accuracy metrics. Green traces correspond to
    raw model output, red traces apply simple KDE-based selection
    based on $64$ samples. Solid traces correspond to evaluation on
    environments containing only circular obstacles, dashed traces to
    environments containing only rectangular
    obstacles. $N_{\text{step}}$ specifies the number of denoising
    steps used in the reverse diffusion process.}
  \label{fig:loc-metrics}
\end{figure}

\begin{figure}[tb]
  \centering
  \includegraphics[width=\linewidth]{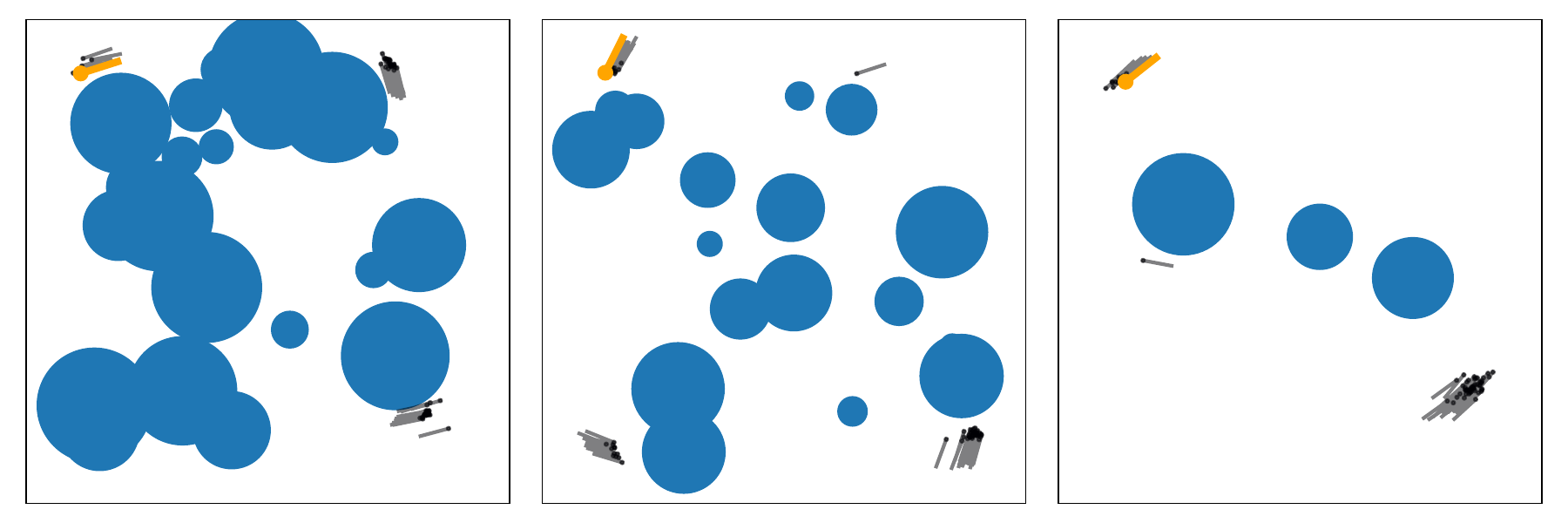}
  \includegraphics[width=\linewidth]{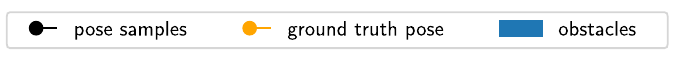}%
  \vspace{-5pt}
  \caption{Typical localization failure behavior corresponding to the
    far right end of the plots from~\cref{fig:loc-metrics}. Note that
    failure scenarios contain symmetry or potentially ambiguous
    elements and model output degrades by gracefully producing
    multimodal output, including samples of the correct mode.}
  \label{fig:loc-failure-cases}
\end{figure}

\textbf{Global localization accuracy.} We first evaluate the global
localization model on a set of $8192$ unseen test examples produced
according to the randomized scenario generation procedure described
in~\cref{sec:dataset-generation}. For each scenario, we sample $64$
poses and compute mean-squared error (MSE) and mean absolute error
(MAE) in position. We additionally perform kernel density (KDE)
estimation with a Gaussian kernel and report MSE and MAE for the
sample with the highest estimated probability. \cref{fig:loc-metrics}
presents MSE and MAE for both the raw model samples as well as the KDE
estimate. Note that with the simple KDE filtering strategy, the model
achieves high localization accuracy of within $2\%$ of the true
position in over~$88\%$ of scenarios. See \cref{fig:loc-failure-cases}
for examples of behavior in failure cases.

\textbf{Performance.} While increasing the number of denoising
iterations $N_{\text{iter}}$ significantly improves the accuracy of
each individual raw sample, note that with the proposed KDE-based
sample selection there is no noticeable drop in sample quality when
reducing $N_{\text{iter}}$ dramatically from 100 to 10. Our
unoptimized implementation can produce 512 samples with 10 denoising
iterations in about 80 \si{ms}.

\begin{figure*}[htb]
  \centering
  \includegraphics[trim=0.7cm 0.2cm 0cm 0cm, width=0.24\linewidth]{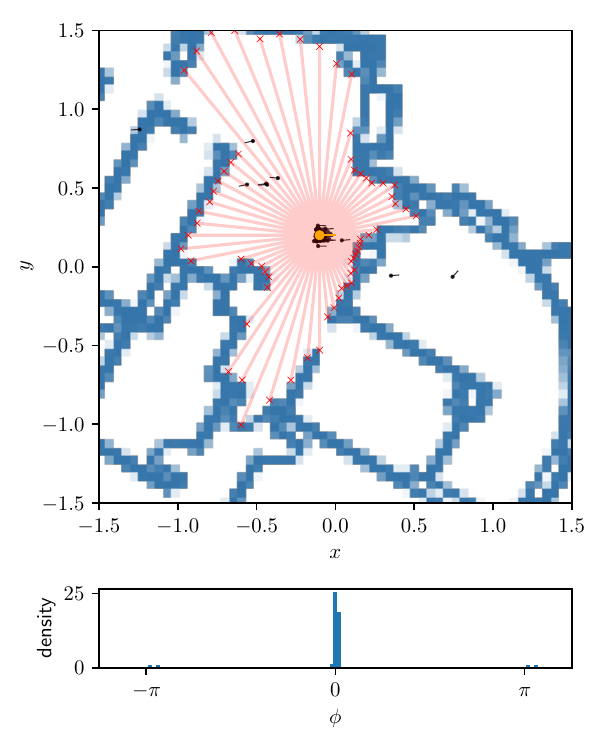}
  \includegraphics[trim=0.7cm 0.2cm 0cm 0cm, width=0.24\linewidth]{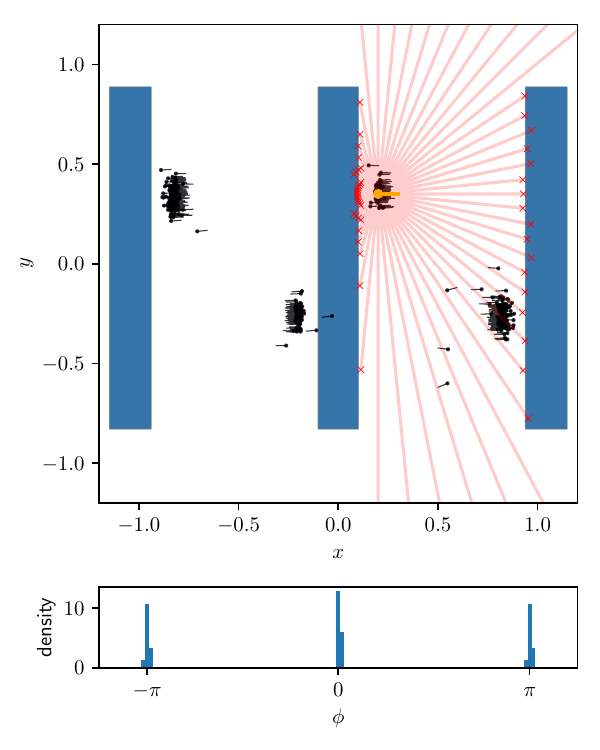}
  \includegraphics[trim=0.7cm 0.2cm 0cm 0cm, width=0.24\linewidth]{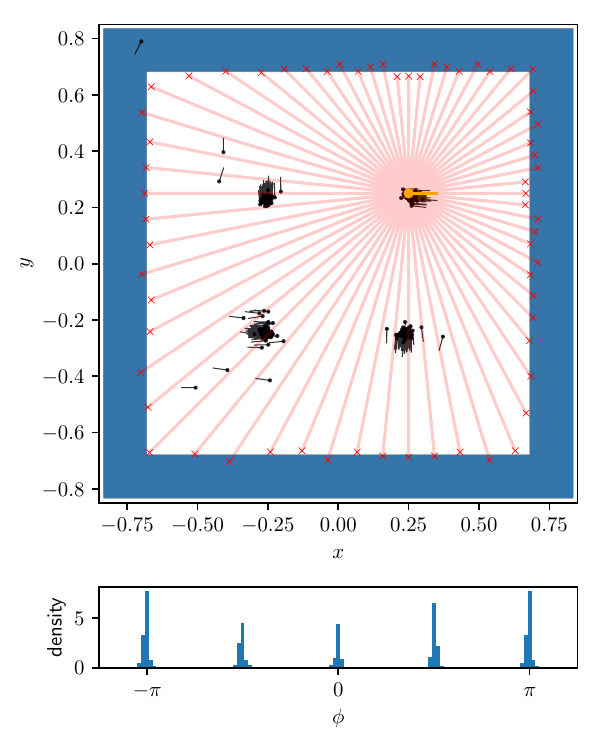}
  \includegraphics[trim=0.7cm 0.2cm 0cm 0cm, width=0.24\linewidth]{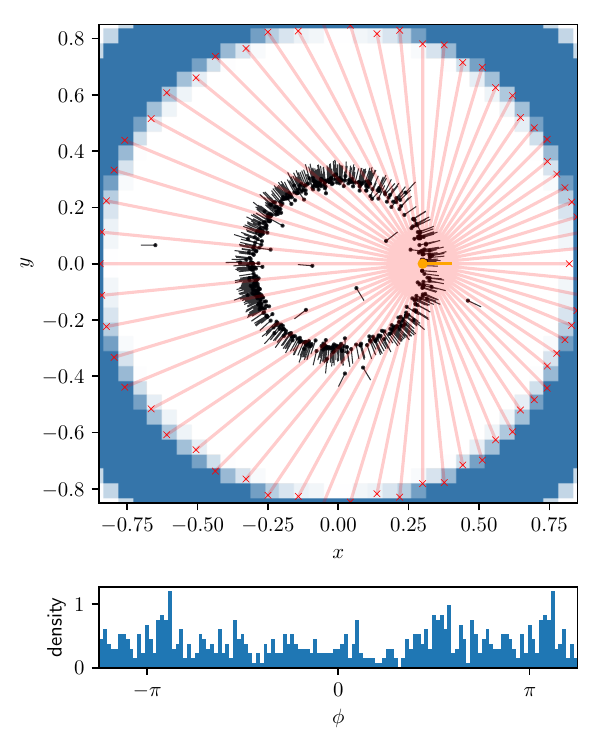}
  \\
  \vspace{-4pt}
  \includegraphics[width=0.5\linewidth]{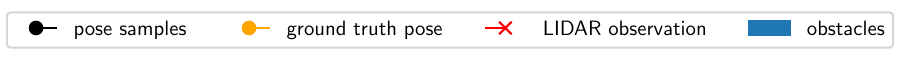}%
  \vspace{-6pt}
  \caption{Localization in hand-designed environments. The LIDAR
    observation is plotted in the coordinate frame of the ground truth
    pose for visualization purposes. The bottom row shows a histogram
    of predicted headings $\phi$ (ground truth heading is zero in each
    scenario). The map in the leftmost scenario is derived from a
    floorplan to illustrate transfer to real-world environments, while
    the three remaining scenarios are designed to illustrate behavior
    in ambiguous scenarios.}
  \label{fig:qualitative-loc-eval}
\end{figure*}

\textbf{Out-of-distribution scenarios.} We implement
rectangular obstacles instead of the circular obstacles
from~\cref{sec:dataset-generation}, according to a similar random
generation procedure. Note that we do not train the model on any such
environments, yet localization accuracy remains comparable to the
circular obstacle cases in the top $80\%$ of scenarios. We refer again
to \cref{fig:loc-metrics} for detailed metrics.

\textbf{Qualitative evaluation.} In \cref{fig:qualitative-loc-eval} we
further analyze a few hand-designed scenarios that are either (i) of
significantly different appearance than the automatically generated
circle/rectangle scenarios used previously, or (ii) exhibit symmetries
that cause the global localization problem to become degenerate,
admitting more than one solution. We find that the model is able to
generalize to unseen classes of obstacle maps such as realistic
floorplans. In scenarios with symmetries, we find that the diffusion
model is able to produce samples that span all possible solution
classes.

\subsection{Joint Localization and Planning}

\begin{table}
  \begin{center}
    \vspace{11.5pt}
    \caption{Joint global localization and planning in synthetic
      environments. When both obstacle types are used, the class of
      each obstacle is selected at random with equal
      probability. ``Success'' implies a well-formed, collision-free
      path and correct global localization within the stated
      tolerance. Length increase reported is total length of model
      solution paths wrt. model-based planner solution and is
      therefore only available for circular obstacle environments.}
    \label{tab:nav-results}
    \begin{tabular}{lccc}\toprule
      & \multicolumn{2}{c}{Success rate [$N_{\text{iter}} =$ 5/15/50] (\%)} & \\ \cmidrule(lr){2-3}
                   & 2\% loc. tol.  & 5\% loc. tol.  & Length incr. (\%) \\\midrule
      Circular     & 81.3/83.4/83.6 & 88.1/89.1/88.6 & 0.5 \\
      Rectangular  & 64.5/66.4/68.9 & 76.8/77.7/79.9 & --- \\
      Both         & 73.0/73.2/73.8 & 82.6/83.2/82.6 & --- \\\bottomrule
    \end{tabular}
  \end{center}
\end{table}

To evaluate the performance of our full model jointly solving the
global localization and path planning problems, we first consider
success rates on $512$ individual in- and out-of-distribution
synthetic examples unseen during training. We then demonstrate how
warm-starting the diffusion process can yield a real-time online
replanning strategy.

\textbf{Synthetic environments.} \cref{tab:nav-results} lists the
results of the evaluation on the synthetic circles and rectangles
datasets. For each scenario, we draw 64 samples with $N_{\text{iter}}$
denoising iterations from the diffusion model and again employ a
Gaussian KDE based sample selection strategy: the sampled trajectory
whose initial pose $T_0^{(0)}$ has the highest probability according
to the KDE result is used. Additionally, we reject any samples that
collide with obstacles.

\textbf{Success criteria.} We define a successful output as having a
localization error of less than certain absolute deviation in each
$x$, $y$, and the heading $\phi$. Note that a 2\% deviation in
position approximately corresponds to the size of one pixel in the
rasterized environment map, so we do not evaluate tolerances lower
than that. We deem a path as colliding if it intrudes by more than
half a pixel width into the obstacles, with this collision check being
performed with respect to the ground truth geometric collection of
obstacles instead of the rasterized environment.

\textbf{Generalization to arbitrary environment maps.} Note that we
implement only support for circular obstacles in our model-based
planner, and therefore, our training data consists only of
environments composed of circular obstacles. Since the model is
conditioned on arbitrary environment maps, in \cref{tab:nav-results}
we evaluate its performance also in environments composed of circular
\textit{and} rectangular obstacles. We observe a modest drop in
success rate when evaluated with the higher 5\% tolerance for
localization error, although there is a more pronounced drop when
considering the tighter 2\% threshold in scenarios with only
rectangular obstacles.

\textbf{Performance.} As shown in \cref{tab:nav-results}, minimal
degradation in success rate and solution quality is observed when
using only 5 denoising iterations. Interestingly, the more
out-of-distribution rectangle-only scenarios appear to benefit more
from a higher number of denoising iterations, while the scenarios
containing circles do not experience a noticeable improvement in
quality even when increasing the number of denoising iterations
dramatically. When reducing the number of denoising iterations to 4 or
lower, the resulting paths are noisy and global localization accuracy
begins to suffer. With 5 denoising iterations, drawing 64 samples from
our unoptimized implementation of the joint global localization and
planning diffusion model takes 140 \si{ms}.

\begin{figure*}[htb]
  \centering
  \includegraphics[trim=0.4cm 0cm 0.4cm 0cm, width=\linewidth]{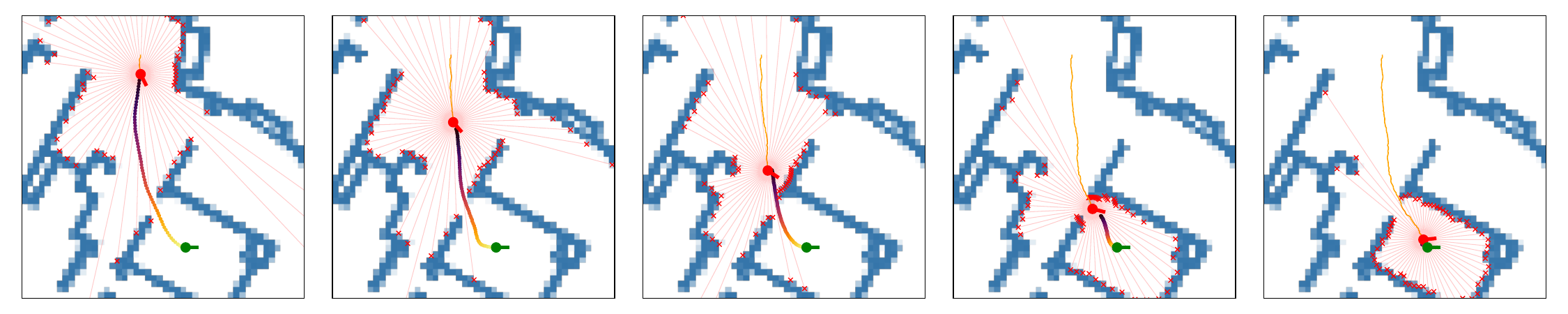}
  \includegraphics[trim=0.4cm 0cm 0.4cm 0cm, width=\linewidth]{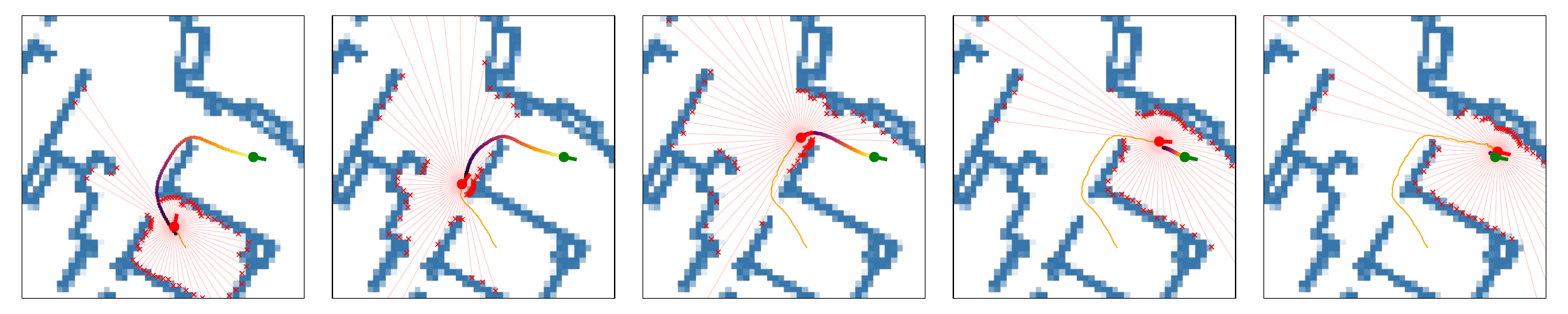}%
  \vspace{-6pt}
  \caption{Navigation with continuous replanning in floorplan
    environment. Fixed goal pose (green), obstacle map (blue) and
    egocentric LIDAR scans serve as conditioning to the diffusion
    navigation model, which produces a globally referenced path
    (multicolored). True vehicle pose is shown in red. Trace of true
    position shown in orange. }
  \label{fig:qualitative-nav-eval}
\end{figure*}

\textbf{Closed loop control.} We deploy the diffusion model for
control of the simple system $\diff x = u(t) \diff t + \mathbf{S}
\diff B_t$ with state $x(t) \in \SE{2}$, control input $u(t) \in
\R^3$, Brownian motion $\diff B_t$, and noise scale $\mathbf{S} =
\text{diag}(\sigma_{xy}^2, \sigma_{xy}^2, \sigma_\phi^2))$. We set
$\sigma_{xy}^2 = 0.1$ and $\sigma_\phi^2 = 0.05$. The control input
$u(t)$ is computed directly from the predicted path by finite
differencing the first two returned poses. We run the full diffusion
process only in the first frame, and warm-start using the previous
solution in subsequent replanning iterations. As mentioned in
\cref{sec:joint-loc-plan} this not only improves computation
efficiency significantly, but also leads to better behavior by
leveraging implicit conditioning on the previous solution to improve
temporal consistency of subsequent plans.
\cref{fig:qualitative-nav-eval} shows two examples of closed-loop
navigation in a realistic environment using this approach.

We also evaluate this replanning and control scheme on the synthetic
environments, with a single denoising iteration in each warm-started
frame. Out of those environments for which global localization in the
first frame succeeds, we find that the model can successfully navigate
the vehicle to the goal pose in 90\% (circular obstacles only), 87\%
(rectangular obstacles only) and 87\% (both obstacle types) of cases.

\textbf{Online replanning performance.} Using warm start, we only
perform a single denoising iteration on a single sample and reuse the
environment map encodings. We can perform such an online replanning
step in~16~\si{ms}, enabling our planning loop to run in real-time at
around~60~\si{Hz}.

\section{Conclusions \& Future Work}

In this work we have developed a diffusion-based model which can
jointly perform global localization on a given map using LIDAR
observations and plan a collision-free path. We demonstrate that the
diffusion framework's powerful distributional modeling abilities
enable the model to gracefully handle degenerate scenarios where
multiple solutions may exist. Furthermore, we find the proposed
conditioning strategies effectively allow our model, trained only on a
narrow set of synthetic examples, to navigate realistic floorplans and
other out-of-distribution scenarios.

While we show that we can already successfully deploy this model for
end-to-end online replanning and control tasks, we identify several
directions for future work. First, we would like to extend our joint
localization and planning model for prediction of multiple timesteps
in order to enable the model to better leverage the coupling of
perception and control through methods like \textit{Diffusion
  Forcing}~\cite{chen2024diffusion}. Next, it would be interesting to
explore the full navigation problem including mapping, instead of
relying on the availability of a map, as well as extensions of our
method to~$\SE{3}$ with camera images instead of LIDAR
scans. Additionally, it would be interesting to investigate the use of
test-time guidance~\cite{song2023loss} instead of or in combination
with the current conditional diffusion model. Finally, we would like
to reconsider the need of an expert planner for dataset generation by
instead training our model for use on vehicles with more complex
dynamics using online reinforcement learning.

We hope that our work on joint global localization and planning can be
a useful stepping stone towards generalizable and robust end-to-end
navigation, enabling the learning of richer behavior than traditional
navigation pipelines that rely on decoupled perception and planning.

\vspace{\fill}

\FloatBarrier

\bibliographystyle{IEEEtran}
\bibliography{references}

\end{document}